\documentclass[]{joyenfra}

\usepackage[toc,page,header]{appendix}

\usepackage{minitoc}
\usepackage{cleveref} 
\usepackage{subcaption}
\usepackage{booktabs}
\usepackage{graphicx}
\usepackage{subcaption}

\newcolumntype{L}[1]{>{\raggedright\arraybackslash}p{#1}}
\newcolumntype{C}[1]{>{\centering\arraybackslash}p{#1}}
\newcolumntype{Y}{>{\raggedright\arraybackslash}X}

\usepackage{fancyhdr} 

\usepackage{natbib}
\usepackage{latexsym}

\usepackage{url}
\usepackage{amssymb}
\usepackage[utf8]{inputenc}
\usepackage{microtype}
\usepackage{booktabs}
\usepackage{pifont} 
\usepackage{multirow}
\usepackage{makecell}
\usepackage{paralist}
\usepackage{xspace}
\usepackage{color}
\usepackage{xcolor}
\usepackage{colortbl}
\usepackage{adjustbox}
\usepackage{hyperref} 
\usepackage[edges]{forest}
\usepackage{tikz} 
\usepackage{caption}
\usepackage{amsfonts}

\hypersetup{
    colorlinks,
    linkcolor={blue!80!black},
    citecolor={blue!80!black},
}
\tikzset{
    root/.style =             {align=center, text width=1cm, rounded corners=3pt, line width=0.3mm, fill=gray!10, draw=gray!80, font=\small},
    demographic/.style =         {align=center, text width=1.8cm, rounded corners=3pt, line width=0.3mm, fill=blue!10, draw=blue!80, font=\footnotesize},
    demographic_work/.style =    {align=center, text width=10cm, rounded corners=3pt, line width=0.3mm, fill=blue!10, draw=blue!0, font=\footnotesize},
    character/.style =         {align=center, text width=1.8cm, rounded corners=3pt, line width=0.3mm, fill=red!10, draw=red!80, font=\footnotesize},
    character_work/.style =    {align=center, text width=10cm, rounded corners=3pt, line width=0.3mm, fill=red!10, draw=red!0, font=\footnotesize},
    personalization/.style =           {align=center, text width=1.8cm, rounded corners=3pt, line width=0.3mm, fill=cyan!10, draw=cyan!80, font=\footnotesize},
    personalization_work/.style =      {align=center, text width=10cm, rounded corners=3pt, line width=0.3mm, fill=cyan!10, draw=cyan!0, font=\footnotesize},
    risk/.style =         {align=center, text width=1.8cm, rounded corners=3pt, line width=0.3mm, fill=orange!10, draw=orange!80, font=\footnotesize},
    risk_work/.style =    {align=center, text width=10cm, rounded corners=3pt, line width=0.3mm, fill=orange!10, draw=orange!0, font=\footnotesize},
}

\usepackage{CJK}

\title{Embodied Operators and Benchmarking: Toward Reusable and Deployable Embodied Intelligence Systems}

\affiliation[1]{AI Infra Team at JDT\quad $^2$Tsinghua University\quad $^3$Tianjin University\quad $^3$Beihang University\quad  
}

\contribution{Full author list in Contributions}

\abstract{
Embodied intelligence systems require not only end-to-end policy models, but also a large number of reusable functional modules that transform multimodal observations, robot states, human demonstrations, and task contexts into structured representations, decisions, trajectories, control references, and system services. This work defines such modules as embodied operators and studies them as independent yet composable units within embodied intelligence pipelines. We first clarify the definition boundary of embodied operators, emphasizing their task semantics, standardized input-output contracts, deployability, reusability, and multi-layer optimizability. We then construct a taxonomy covering five major categories: detection and segmentation operators, spatial localization and 3D understanding operators, hand motion recovery operators, embodied foundation models and task-decision operators, and planning, control, and system support operators. For each category, we summarize representative functions, technical paradigms, application roles, and practical limitations in manipulation-oriented embodied systems. Beyond operator taxonomy, we further propose a multi-dimensional benchmark framework for embodied operators. Unlike conventional single-model evaluation, the proposed benchmark considers correctness, end-to-end efficiency, resource usage, temporal stability, portability, interface compatibility, deployment reliability, and downstream task utility. We also discuss operator acceleration from a workflow-level perspective and analyze open challenges in operator composition, data standardization, world models, VLA safety, edge deployment, and real-world application value. Overall, this work argues that embodied operators should be optimized and evaluated as holistic deployable components rather than isolated neural networks, providing a foundation for reusable, scalable, and verifiable embodied intelligence systems.

}

\date{\today}
\correspondence{Junwu Xiong \email{xiongjunwu.1@jd.com}}

\begin{document}

\maketitle


\section{Introduction}

Embodied intelligence aims to enable physical or simulated robotic agents to perceive their environments, understand task contexts, and perform interactions through embodied actuation. Recent advances in vision-language-action models, robot learning datasets, simulation platforms, and open-source robot learning systems have substantially enhanced the capabilities of embodied agents \cite{liu2025aligning,kim24openvla,team2024octo,black2024pi_0,bjorck2025gr00t,o2024open,khazatsky2024droid,cadene2026lerobot}. However, high-quality embodied intelligence systems cannot rely solely on end-to-end policy models. In practical pipelines for data collection, demonstration understanding, scene reconstruction, robot learning, task decision-making, and robot execution, numerous reusable functional modules are required to transform raw multimodal signals and intermediate system states into structured representations, decisions, and executable outputs. In this work, we refer to such reusable functional modules as embodied operators.

Different from low-level tensor computation primitives, an embodied operator is a deployable functional unit \cite{mayoral2022robotcore} with explicit task semantics, standardized input-output interfaces, and measurable performance characteristics \cite{mayoral2024robotperf}. In embodied data collection and robot execution, these operators play a central role in converting RGB images, depth observations, videos, point clouds, robot states, language instructions, and human demonstrations into usable annotations, representations, plans, and control signals. Typical outputs include hand bounding boxes, hand masks, object poses, camera trajectories, depth maps, reconstructed scenes, hand keypoints, hand motion trajectories, action proposals, grasp poses, collision-free trajectories, controller references, and system messages. These outputs provide essential support for downstream tasks such as demonstration parsing, manipulation policy learning, imitation learning, simulation asset construction, task planning, and closed-loop robot execution.

This white paper focuses on a set of core embodied operators that are frequently used in manipulation-oriented embodied intelligence pipelines. We first clarify the definition and boundary of embodied operators and construct a taxonomy for operators in embodied intelligence workflows. We then analyze hand-object detection and segmentation operators, which identify hands, objects, and interaction regions from visual observations and provide essential cues for human demonstration understanding and interaction-centered data annotation. This discussion covers their methodological taxonomy, evaluation criteria, and quality-preserving acceleration strategies. Next, we examine spatial localization and 3D understanding operators, including SLAM-based localization, object-level pose estimation, depth prediction, scene reconstruction, and dense mapping. We further discuss hand motion recovery operators, focusing on their problem formulation, input-output formats, representative model paradigms, and key limitations in reconstruction quality and efficiency. In addition, we review embodied foundation models and task-decision operators, including vision-language models, vision-language-action models, and world models. We also introduce planning, control, and system support operators, including grasp planning, trajectory planning, collision checking, ROS~2 adaptation, data transfer, and heterogeneous scheduling. Finally, we present a multi-dimensional benchmark for evaluating embodied operators from both functional and system-level perspectives.

The motivation for studying embodied operators as independent yet composable functional units is twofold. First, embodied intelligence systems require reliable intermediate representations and executable outputs, rather than only final policy predictions. Errors in hand localization, mask quality, object pose estimation, depth prediction, hand motion recovery, action generation, trajectory planning, collision checking, or system synchronization can directly propagate to downstream learning or execution modules and degrade task performance. Second, practical deployment is constrained not only by prediction accuracy, but also by runtime efficiency, memory pressure, temporal stability, coordinate and timestamp consistency, cross-hardware compatibility, failure recovery, and integration cost. Therefore, embodied operators should be evaluated from a system-level perspective that jointly considers representation quality, computational overhead, execution reliability, and downstream usability.

The central claim of this work is that embodied operators should be optimized and evaluated as holistic functional units rather than isolated neural networks. A useful operator should not only achieve strong benchmark accuracy, but also provide standardized interfaces, stable outputs, controllable failure behavior, efficient runtime performance, and clear compatibility with upstream and downstream modules. This perspective is particularly important for building reusable embodied operator libraries, in which perception, localization, reconstruction, motion recovery, task decision, planning, control, communication, and runtime support modules must be integrated into reproducible and scalable workflows.

The main contributions of this work are summarized as follows:
\begin{itemize}
    \item We define the concept and boundary of embodied operators and construct a taxonomy that connects visual perception, spatial understanding, human and action understanding, embodied foundation models, task decision-making, planning, control, and system support.

    \item We systematically review hand-object detection and segmentation operators, including their task formulation, representative technical paradigms, evaluation metrics, and quality-preserving acceleration strategies.

    \item We analyze spatial localization and 3D understanding operators, covering SLAM-based localization, object-level pose estimation, depth prediction, scene reconstruction, dense mapping, and their limitations in practical embodied intelligence scenarios.

    \item We discuss hand motion recovery operators from the perspectives of problem definition, input-output representations, model paradigms, reconstruction quality, efficiency, and open challenges.

    \item We review embodied foundation models and task-decision operators, including vision-language models, vision-language-action models, and world models, and discuss their roles in semantic understanding, action generation, and future-state prediction.

    \item We analyze planning, control, and system support operators, including grasp planning, trajectory planning, collision checking, controller execution, ROS~2 adaptation, data transfer, and heterogeneous scheduling.

    \item We formulate a multi-dimensional evaluation perspective for embodied operators, emphasizing prediction quality, runtime efficiency, temporal robustness, interface standardization, deployment compatibility, execution reliability, and downstream task utility.
\end{itemize}

\section{Background and Taxonomy of Embodied Operators}

\subsection{Definition Boundary}

In this work, an embodied operator refers to a reusable computational module that performs a well-defined function within an embodied intelligence system. It is broader than a low-level deep learning operator or hardware kernel, as its role is not limited to numerical computation. Instead, it transforms sensory inputs, spatial information, human behaviors, task contexts, robot states, or intermediate system states into representations, decisions, trajectories, control references, or system services that can be consumed by downstream embodied modules. Therefore, an embodied operator carries both computational semantics and task semantics. Its value lies not only in producing accurate outputs, but also in providing stable interfaces, predictable runtime behavior, and reliable support for subsequent perception, planning, learning, control, and deployment processes.

The boundary of an embodied operator can be characterized by five key properties. First, it has functional independence, meaning that it completes a clearly defined embodied task, such as segmentation, pose recovery, depth estimation, action generation, grasp planning, trajectory planning, collision checking, message adaptation, or heterogeneous scheduling. Second, it has an explicit input-output contract, allowing other modules to consume its outputs without relying on internal implementation details. For operators involving geometric or temporal information, this contract may further specify coordinate frames, timestamps, units, confidence scores, constraints, and failure states. Third, it is reusable across different robot systems, data production workflows, and application scenarios. Fourth, it is deployable in practical forms, such as a service, SDK, ROS~2 node \cite{macenski2022ros2}, plugin, offline processing tool, or runtime component \cite{mayoral2022robotcore}. Fifth, it is optimizable across multiple layers, including model architecture, computation graph, low-level operators, hardware kernels, data movement, serialization, and runtime scheduling.

\begin{table}[t]
\centering
\caption{Representative categories of embodied operators and example modules.}
\label{tab:operator_taxonomy}
\renewcommand{\arraystretch}{1.25}
\begin{tabular}{p{0.22\linewidth}p{0.34\linewidth}p{0.34\linewidth}}
\toprule
\textbf{Category} & \textbf{Representative Operators} & \textbf{Functional Role} \\
\midrule
Visual perception &
SAM2, Grounding DINO, YOLO-World, CoTracker, hand keypoint detectors &
Extract object masks, target locations, tracks, keypoints, and visual interaction regions from images or videos. \\
\midrule
Spatial localization and 3D understanding &
FoundationPose, ORB-SLAM3, Depth Anything, DUSt3R/MASt3R, VGGT, 3D reconstruction &
Estimate object poses, camera motion, depth, scene geometry, and spatial constraints for manipulation and navigation. \\
\midrule
Human and action understanding &
Dyn-HaMR, HaMeR, WiLoR~\cite{potamias2025wilorendtoend3dhand}, HaWoR~\cite{zhang2025haworworldspacehandmotion}, MANO~\cite{MANO}/SMPL reconstruction, action retargeting &
Convert human demonstrations into structured motion, pose, trajectory, and interaction representations. \\
\midrule
Embodied foundation model and task decision &
Qwen-VL, RT-2, OpenVLA, $\pi_0$, VLA models, world models &
Map visual observations, language instructions, robot states, and historical context into plans, actions, or predicted future states. \\
\midrule
Planning, control, and system support &
Contact-GraspNet, AnyGrasp, MoveIt 2, cuRobo, ROS 2/NITROS, heterogeneous scheduling &
Transform perception and decision outputs into executable robot behaviors while coordinating communication and computing resources. \\
\bottomrule
\end{tabular}
\end{table}
\FloatBarrier

\subsection{Taxonomy of Embodied Operators}

From the perspective of embodied intelligence workflows, embodied operators can be organized into five representative categories. The first category is visual perception operators, including object detection, image and video segmentation, target tracking, and keypoint detection. These operators transform raw visual observations into semantic cues, spatial regions, and interaction-related information for downstream modules. The second category is spatial localization and 3D understanding operators, including depth estimation, 6D pose estimation, SLAM, point cloud processing, 3D reconstruction, occupancy prediction, and visual odometry. These operators enable robots to infer object locations, camera or robot motion, and the geometric structure of the surrounding environment.

The third category is human and action understanding operators, including human pose estimation, hand reconstruction, motion recovery, action recognition, action segmentation, and action retargeting. These operators are essential for converting human demonstrations into structured trajectories and reusable robot learning data. The fourth category is embodied foundation model and task-decision operators, including vision-language models, vision-language-action models, world models, task planners, and action generators. These operators connect high-level instruction understanding, scene reasoning, future-state prediction, and robot action generation. The fifth category is planning, control, and system support operators, including grasp planning, trajectory planning, collision checking, controller execution, ROS~2 message adaptation, data transfer, and heterogeneous scheduling. These operators convert perception and task-decision outputs into feasible robot behaviors while coordinating the computational and communication resources required for execution.

Although the fourth and fifth categories are closely related, they play distinct functional roles. Embodied foundation models and task-decision operators mainly determine what action or task should be performed, whereas planning, control, and system support operators determine how the action can be executed under kinematic, dynamic, collision, communication, and runtime constraints.

These categories are not isolated, but are typically composed into complete embodied workflows \cite{o2024open,cadene2026lerobot}. In a robotic manipulation pipeline, an object detection operator may first identify the target, a segmentation operator may extract its mask, a pose estimation operator may recover its 6D pose, and a VLA model or task planner may generate an action proposal. Subsequently, grasp planning, trajectory planning, collision checking, and control operators convert the proposal into executable robot motion. Meanwhile, ROS~2 adaptation \cite{macenski2022ros2}, data transfer, and heterogeneous scheduling operators coordinate communication and computation across the full perception-action pipeline. In embodied data production, a hand motion recovery operator may convert human operation videos into structured hand trajectories, which can then be cleaned, retargeted, and used for imitation learning or VLA fine-tuning. Therefore, the construction of an embodied operator library should go beyond accumulating individual modules; it should also standardize their interfaces, optimize data exchange among them, and evaluate their contributions to complete task workflows.

\section{Detection and Segmentation Operators}

Detection and segmentation constitute fundamental perceptual capabilities in embodied intelligence systems. During robotic manipulation, hands and objects jointly provide critical visual cues regarding action intent, manipulation targets, physical contact, and environmental states. In particular, hands typically reflect the agent's manipulation intent and motion patterns, whereas objects correspond to manipulation targets, tools, or environmental constraints. Detection and segmentation operators are therefore required not only to localize hands and objects, but also to provide spatial priors for downstream tasks such as hand-object interaction understanding \cite{carfi2021hand}, contact inference, motion reconstruction, learning from demonstration, and robot control.Detection and segmentation differ in terms of spatial granularity. Detection typically outputs bounding boxes for hands, objects, or tools, providing coarse regional priors for subsequent segmentation, tracking, and reconstruction. Segmentation, by contrast, produces pixel-level masks to recover accurate object boundaries and separate hands and objects from the background, occluded regions, or other interacting entities. Accordingly, detection emphasizes object discovery and efficient localization, whereas segmentation emphasizes boundary accuracy and spatial parsing capability.

Detection and segmentation methods for embodied intelligence can be broadly divided into two categories: conventional detection and segmentation methods, and next-generation detection and segmentation methods. Conventional methods typically rely on predefined categories and task-specific datasets for supervised training, making them suitable for tasks with clearly defined category spaces and relatively stable scenes. Next-generation methods, in contrast, emphasize stronger generality, interactivity, and openness. For example, they can perform open-vocabulary detection through textual prompts, or flexible segmentation through prompts such as points, boxes, masks, and other forms of guidance, thereby better adapting to unknown objects and task variations in open embodied environments.

\subsection{Conventional Detection and Segmentation Methods}

Conventional detection and segmentation methods typically regard hands, tools, and manipulated objects as predefined target categories, and train detection or segmentation models through supervised learning. For detection tasks, lightweight hand detection methods such as MediaPipe Hand Landmarker \cite{lugaresi2019mediapipe} can estimate hand keypoints and handedness in real time. General-purpose object detection frameworks such as YOLOv7 \cite{wang2023yolov7} and DINO \cite{zhang2022dino} can be used to localize entities such as hands, objects, tools, and robotic end-effectors. These methods are fast, easy to deploy, and perform reliably in fixed scenes and known-category tasks.

For segmentation tasks, conventional semantic segmentation, instance segmentation, and hand-object segmentation methods are typically trained on task-specific datasets to generate pixel-level masks of hands or objects. Among them, interaction-aware joint hand-object segmentation methods further explicitly model the relationship between hands and active objects. For example, ORMNet \cite{su2024ormnet} proposes object-centric relation modeling, CaRe-Ego \cite{su2025care} introduces contact-aware relation modeling, and InterFormer \cite{su2026interaction} employs interaction-aware queries and co-occurrence consistency to improve hand-object parsing. Compared with generic segmentation models, these methods are better suited for handling hand-object occlusion, contact boundaries, and interactive target identification.

\subsection{Next-Generation Detection and Segmentation Methods}

The key characteristic of next-generation detection and segmentation methods is the transition from fixed categories and closed scenes toward more general and flexible object perception. Their novelty is mainly reflected in two aspects. On the one hand, detection models increasingly support open-vocabulary object localization, enabling the detection of hands, tools, and task-relevant objects through natural-language prompts. On the other hand, segmentation models increasingly support promptable segmentation, generating target masks based on prompts such as points, boxes, text, masks, or visual examples.

For detection, Grounding DINO \cite{liu2024grounding} supports text-conditioned object localization and can detect hands, tools, and interacting objects according to prompt words. Grounding DINO 1.5 \cite{ren2024grounding} further improves the deployment efficiency of open-vocabulary detection. YOLO-World \cite{cheng2024yolo} combines open-vocabulary detection with real-time inference, making it suitable for robotic applications that require rapid responses. These methods are no longer limited to a fixed category set; instead, the same model can flexibly localize different targets according to task requirements, making them more suitable for embodied perception in open environments.

For segmentation, models such as SAM \cite{kirillov2023segment} represents the development of promptable segmentation. The key innovation of SAM lies in its general promptable segmentation paradigm, which can generate target masks from point, box, or mask prompts. SAM2 \cite{ravi2025sam2} extends this paradigm to video scenarios, enabling masks to be propagated and tracked over long sequences. SAM3 \cite{carion2025sam} further supports concept-level detection, segmentation, and tracking based on text, visual examples, or hybrid prompts . In embodied scenarios, these models can generate hand and object masks using hand boxes, object boxes, keypoints, or textual descriptions, thereby reducing manual annotation costs and improving cross-scene adaptability.

\subsection{Limitations}

Although existing detection and segmentation operators have made substantial progress in general object localization, open-vocabulary detection, and promptable segmentation, their robustness in embodied intelligence scenarios still requires further improvement. Robotic manipulation often involves severe occlusion, fast motion, hand-object contact, viewpoint changes, and background clutter. Existing methods may still suffer from missed detections, false positives, or unstable mask boundaries when handling mutual occlusion between hands and objects, fine-grained contact boundaries, and previously unseen scenes. Therefore, improving model stability under complex interactions and cross-scene transfer remains an important direction for future detection and segmentation operators.

In addition, real robotic systems impose stricter real-time requirements on perception modules. Many high-accuracy open-vocabulary detection and promptable segmentation models incur substantial computational costs, and direct deployment in real-robot closed-loop control systems may fail to meet low-latency requirements. In large-scale data collection and offline data processing scenarios, the speed of detection and segmentation also directly affects the efficiency of data cleaning, annotation generation, and demonstration data construction. Future methods therefore need to further improve inference speed and engineering deployability while maintaining high spatial accuracy and open-world generalization capability.

Overall, detection and segmentation operators must balance accuracy, speed, and generalization. High-accuracy models typically provide more stable boundaries and stronger adaptability to open scenes, but at the cost of higher computation. Lightweight models are more suitable for real-time control and large-scale data processing, but may exhibit degraded performance under occlusion, unseen objects, and complex interaction scenarios. Selecting appropriate model scales and operator combinations according to the requirements of embodied tasks is therefore a key consideration in practical system design.

\section{Spatial Localization and 3D Understanding}

Spatial localization and 3D understanding provide the geometric foundation for embodied operators, enabling agents to estimate their own motion, object poses, and scene structures, thereby supporting downstream navigation and manipulation tasks.

\subsection{SLAM as a Localization Operator}

SLAM, or simultaneous localization and mapping, refers to the process by which an agent simultaneously estimates its own pose and constructs a map of the environment from sensor observations in an unknown environment. As a localization operator in embodied systems, SLAM provides a continuous spatial reference for navigation, path planning, scene reconstruction, and manipulation tasks.

In recent years, SLAM research has rapidly shifted from classical optimization-based pipelines and early learning-based methods toward approaches based on Transformers and neural scene representations. ORB-SLAM3 \cite{campos2021orb} is a widely used classical visual SLAM system and remains a strong baseline among feature-based SLAM methods.

Recent methods have increasingly explored SLAM systems based on Gaussian splatting and reconstruction priors. Representative examples include SplaTAM \cite{keetha2024splatamsplattrack}, which introduces 3D Gaussian Splatting into dense RGB-D SLAM to achieve high-fidelity real-time reconstruction, and MASt3R-SLAM \cite{murai2025mast3rslamrealtimedenseslam}, which leverages strong geometric reconstruction priors from visual foundation models to achieve robust tracking and dense mapping in complex, real-world open scenes.

\subsection{Object-Level Pose Estimation Operators}

Object pose estimation aims to recover the 6D position and orientation of individual objects from RGB or RGB-D observations.

FoundationPose \cite{wen2024foundationpose} is a representative CAD-model-based method. Given the CAD model of a known object, it performs category-agnostic 6D pose estimation and tracking. It can provide accurate spatial localization of rigid objects for manipulation tasks such as grasping, assembly, and tool use, and is particularly suitable for scenarios where object geometry is known.

Furthermore, BundleSDF \cite{wen2023bundlesdfneural6doftracking} provides a complementary capability. By jointly optimizing camera tracking and dense surface reconstruction, it recovers an implicit signed distance function (SDF) representation of previously unseen objects. Based on RGB-D input, this method can reconstruct object-level geometry and progressively infer a pseudo-CAD model, thereby supporting downstream pose estimation without requiring an explicit CAD model.

By combining CAD-based methods for known objects, such as FoundationPose, with reconstruction-based methods for unknown objects, such as BundleSDF, a unified pose estimation pipeline can be constructed to support both known instances and novel object scenarios. This hybrid formulation improves the generality of spatial localization in robotic manipulation and embodied interaction tasks.

\subsection{Scene Understanding and 3D Reconstruction Operators}

Scene-level 3D understanding focuses on recovering the geometric structure and spatial relationships of the environment from visual observations, and serves as an important foundation for navigation, manipulation, interaction, and environmental modeling in embodied intelligence systems. According to task hierarchy, related methods can be broadly divided into two categories. The first category concerns fundamental scene understanding capabilities, such as depth estimation, surface normal prediction, and 3D point map recovery, which mainly extract local or dense geometric priors from a single frame or a small number of images. The second category concerns higher-level 3D reconstruction tasks, which aim to integrate multi-view observations into a unified, consistent 3D scene representation that can support robotic decision-making. Compared with classical reconstruction pipelines that rely on sparse feature matching and explicit geometric optimization \cite{schonberger2016structure}, modern methods increasingly emphasize dense perception, learned geometric priors, and neural scene representations.

\paragraph{Fundamental Scene Understanding Capabilities.}
Fundamental scene understanding capabilities mainly include depth prediction, metric geometry estimation, surface normal prediction, and 3D point structure recovery. Among them, depth estimation is a core component of 3D understanding, as it converts RGB images into relative or metric geometric information and provides robots with basic priors about scene layout, object distance, and traversable space. Recent learning-based methods leverage large-scale pretraining and multi-view supervision to substantially improve the robustness of dense depth prediction in textureless regions, complex lighting conditions, and cross-domain scenarios. Depth Anything \cite{lin2025depth3recoveringvisual} is a representative recent method that demonstrates strong generalization capability for monocular depth estimation across diverse environments.

Beyond Depth Anything, methods such as Metric3D \cite{yin2023metric}, UniDepth \cite{piccinelli2024unidepth, piccinelli2025unidepthv2}, and MoGe \cite{wang2025moge, wang2025moge2} further extend the capability boundary of monocular geometric understanding. Metric3D targets zero-shot metric depth estimation and surface normal prediction, enabling the recovery of scale-aware geometric information across different cameras and scene conditions. UniDepth further emphasizes predicting metric depth or 3D point structures from a single image, allowing monocular observations to provide more direct 3D spatial priors. MoGe focuses on open-domain monocular geometry estimation and can recover not only depth, but also 3D point maps, surface normals, and camera field-of-view-related geometric attributes. For embodied intelligence systems, these fundamental geometric operators transform 2D visual observations into 3D cues for spatial reasoning, providing low-level support for object localization, reachability analysis, motion planning, and interaction understanding.

\paragraph{High-Level 3D Reconstruction Tasks.}
High-level 3D reconstruction tasks aim to integrate single-view or multi-view observations into a unified and consistent 3D scene representation, such as depth maps, point clouds, implicit geometric fields, or explicit Gaussian representations, thereby enabling holistic understanding of scene structure, free space, and object layout. Unlike fundamental depth estimation, which mainly focuses on single-frame or local geometry, 3D reconstruction places greater emphasis on cross-view consistency, camera parameter estimation, global geometric fusion, and structured representations for downstream tasks.

VGGT \cite{wang2025vggtvisualgeometrygrounded} is a representative recent feed-forward visual geometry model that can directly predict camera parameters, depth, and dense 3D point structures from single-view or multi-view inputs. Compared with traditional reconstruction pipelines that rely on explicit optimization such as bundle adjustment, VGGT provides a unified and efficient solution for dense 3D reconstruction. Related methods such as DUSt3R \cite{wang2024dust3rgeometric3dvision} and MASt3R \cite{leroy2024groundingimagematching3d} also emphasize feed-forward multi-view 3D reconstruction. DUSt3R directly models dense correspondences between arbitrary image pairs through Transformers and recovers 3D point structures, while MASt3R further introduces stronger cross-view matching capabilities, making it more suitable for multi-image reconstruction and large-scale scene modeling. These methods reduce the dependence on complex feature matching, camera calibration, and iterative optimization in traditional SfM/MVS pipelines \cite{schonberger2016structure}, offering new possibilities for robots to rapidly construct 3D scene representations from unstructured visual inputs.

In addition, reconstruction methods based on neural representations and explicit Gaussian representations provide new technical routes for high-fidelity scene modeling. NeRF-based methods represent scene geometry and appearance through implicit neural radiance fields; for example, NeRFPrior \cite{zhang2025nerfprior} uses NeRF as a geometric and color prior to improve the surface reconstruction quality of multi-view indoor scenes. 3D Gaussian Splatting-based methods represent scenes with explicit 3D Gaussians, improving representation and rendering efficiency while maintaining high rendering quality. For instance, ObjectGS \cite{zhu2025objectgs} further introduces object-level modeling, enabling Gaussian representations to support not only scene reconstruction but also object-level scene understanding. For embodied intelligence systems, these methods can serve as complementary high-fidelity 3D scene representations, supporting interaction space analysis, viewpoint selection, and long-term scene memory. Meanwhile, recent SAM3D \cite{chen2026sam}  further extends the promptable segmentation paradigm to 3D object understanding, recovering object geometry, appearance, and spatial layout from visual inputs. For embodied intelligence systems, such methods can provide richer 3D representations for interaction space analysis, viewpoint selection, and long-term scene memory.

\subsection{Limitations}

Although spatial localization and 3D understanding have made significant progress in recent years, their practical application in embodied intelligence systems still faces challenges in terms of fidelity and efficiency. On the one hand, common factors in robotic scenarios, such as occlusion, dynamic objects, reflective materials, rapid viewpoint changes, and hand-object contact, can lead to missing geometry, surface noise, or unstable pose estimation. These factors make it difficult for existing methods to continuously obtain high-fidelity, scale-consistent, and structurally complete 3D representations in open environments.

On the other hand, many high-accuracy 3D understanding methods rely on complex network inference, cross-view matching, or iterative optimization, resulting in high computational costs that make it difficult to directly meet the low-latency and online-response requirements of real robotic systems. In large-scale data collection and offline processing scenarios, inference speed also affects the efficiency of data cleaning, trajectory annotation, and scene modeling.

Therefore, spatial localization and 3D understanding operators need to balance accuracy, speed, and fidelity. High-fidelity methods are more suitable for fine-grained manipulation and accurate spatial reasoning, but they incur higher deployment costs. Lightweight methods are more suitable for real-time control and large-scale data processing, but may sacrifice geometric detail and cross-scene robustness. How to adaptively select 3D representations, model scales, and computational strategies according to the requirements of embodied tasks remains an open question worthy of further investigation.

\section{Hand Motion Recovery Operators}

Hand Motion Recovery (HMR) operators are fundamental components in embodied data generation pipelines. Their goal is to recover the three-dimensional motion and geometry of human hands from visual observations, and to produce structured representations that can be directly used for robot learning, simulation, and downstream embodied intelligence tasks. As a bridge between raw visual demonstrations and high-level manipulation semantics, HMR operators play a key role in action understanding, grasp synthesis, imitation learning, and hand-object interaction modeling \cite{carfi2021hand}.

\subsection{Problem Definition}

Hand motion recovery aims to infer the three-dimensional state of the human hand from visual observations. Formally, this task can be viewed as learning a mapping from two-dimensional visual signals to structured three-dimensional geometric and kinematic representations. In embodied data generation pipelines, HMR operators are typically applied after video preprocessing, providing foundational inputs for subsequent trajectory reconstruction, manipulation understanding, policy learning, and robot motion retargeting.

Depending on the data acquisition setup, the input to hand motion recovery can take multiple forms. The most common setting is monocular RGB images or videos, with data sources including Internet videos, third-person manipulation videos, and egocentric recordings captured by wearable cameras. In addition to RGB inputs, depth information, hand keypoints, detection boxes, or segmentation masks can be introduced as priors to alleviate scale ambiguity and improve geometric accuracy. Moreover, multi-view image sequences can obtain more accurate 3D geometry through triangulation, and are often used for high-quality annotation and benchmark evaluation. Motion capture (MoCap) systems are also commonly used for high-precision data collection, providing reliable supervision signals for model training and evaluation.

The output of an HMR operator depends on the requirements of downstream tasks and can be represented at multiple levels of abstraction. Low-level outputs usually include single-frame 3D hand keypoints and parameters of a parametric hand model, such as the pose and shape parameters of MANO \cite{MANO}. Higher-level outputs may further include hand-object contact states, global hand trajectories, and temporally optimized continuous hand motion representations. These structured outputs enable visual demonstrations to be converted into motion data that are better suited for robot learning and simulation.

\subsection{Model Paradigms}

Hand motion recovery methods have evolved from early sparse keypoint estimation to parametric hand mesh reconstruction, and have further progressed toward large-capacity Transformer-based models and temporally aware models. Early systems mainly focused on localizing hand joints in images, whereas modern methods aim to recover dense 3D hand geometry, global pose, and temporally consistent motion trajectories. This shift is particularly important for embodied data generation, because the recovered hand representation needs to be not only visually accurate, but also geometrically complete, temporally stable, and suitable for downstream robot motion retargeting and interaction modeling.

\subsection{Methods Based on the MANO Hand Model}

Methods based on parametric hand models typically adopt MANO \cite{MANO} as a structured hand representation, and recover the 3D hand mesh by regressing MANO pose and shape parameters from visual observations. Such methods compress complex hand geometry into a low-dimensional parameter space, and are therefore widely used in 3D hand reconstruction and motion recovery.

Representative hand-specific methods include HandOccNet \cite{park2022handoccnetocclusionrobust3dhand} and AMVUR \cite{jiang2023probabilisticattentionmodelocclusionaware}. HandOccNet targets occlusion-robust 3D hand mesh estimation, aiming to recover plausible hand meshes when fingers or palm regions are occluded by objects or affected by self-occlusion. AMVUR alleviates ambiguity in monocular 3D hand reconstruction by introducing uncertainty-aware mesh regression, which helps model reconstruction uncertainty under depth ambiguity, occlusion, and incomplete visual evidence.

Although these methods improve the robustness of MANO-based hand reconstruction, they are still constrained by limited model capacity, restricted training data scale, and frame-by-frame prediction paradigms. In complex manipulation videos, single-frame reconstruction is easily affected by occlusion, motion blur, and viewpoint variation, leading to unstable finger poses or discontinuous trajectories. These limitations have motivated the development of larger-scale Transformer-based models and temporal hand motion recovery methods.

Recent HMR methods have further moved toward more robust, scalable, and temporally consistent reconstruction, focusing on challenges posed by large-scale Internet videos, egocentric demonstrations, third-person manipulation videos, severe occlusion, motion blur, and camera motion.

HaMeR \cite{pavlakos2023reconstructinghands3dtransformers} is a representative large-scale Transformer-based method for monocular hand mesh recovery. It uses a high-capacity Vision Transformer and is trained on a mixture of 2D and 3D hand datasets, directly predicting MANO parameters from cropped hand images. By scaling up model capacity and training data, HaMeR significantly improves hand reconstruction robustness in in-the-wild scenarios, and is therefore often used as a front-end module for generating hand pseudo-labels from large-scale visual demonstrations.

WiLoR \cite{potamias2025wilorendtoend3dhand} improves the practicality of hand recovery in in-the-wild videos by integrating hand localization and 3D reconstruction. Unlike methods that assume clean hand crops as input, WiLoR focuses more on complex inputs in real videos, such as incomplete hand regions, multiple hands in the frame, or hands mixed with complex backgrounds and manipulated objects.

HaWoR \cite{zhang2025haworworldspacehandmotion} further extends hand recovery to motion reconstruction in world space, especially for egocentric videos. It not only estimates hand pose in the camera coordinate system, but also combines camera-space hand reconstruction, camera trajectory estimation, and motion completion to recover continuous hand trajectories in a stable world coordinate system. Compared with methods that only output local hand meshes, world-coordinate hand trajectories are more suitable for robot imitation learning, manipulation trajectory analysis, and cross-agent motion transfer.

These studies are important for embodied data generation. Accurately recovered hand poses and trajectories can provide strong data priors for robot manipulation learning, reduce manual annotation costs, and enable the automated construction of large-scale embodied datasets.

\subsection{Limitations}

Despite significant recent progress, current HMR operators still face two major challenges in embodied data generation: reconstruction quality and processing efficiency.

First, reconstruction quality still falls short of the requirements of complex real-world manipulation scenarios. Although modern methods can recover plausible hand meshes in many in-the-wild scenarios, they remain unstable under severe occlusion, motion blur, uncommon viewpoints, and ambiguous hand-object interactions. In manipulation videos, fingers are frequently occluded by objects, making it difficult to accurately estimate finger joints, palm poses, and contact regions. In addition, many methods still suffer from temporal inconsistency, such as finger jitter, local pose jumps, or global hand trajectory drift. These errors directly affect downstream tasks, including contact estimation, motion retargeting, grasp synthesis, and robot trajectory reproduction.

Second, efficiency is a key bottleneck in large-scale embodied data collection. High-quality hand recovery pipelines usually rely on large-capacity Transformer backbones and need to combine multiple modules, including hand detection, tracking, temporal smoothing, camera pose estimation, or world-coordinate reconstruction, resulting in high computational cost and inference latency. For large-scale video datasets, HMR operators must not only process long sequences automatically, but also reduce computational overhead while maintaining accuracy and temporal consistency. Therefore, future methods need to better balance reconstruction accuracy, temporal stability, and inference speed, so that hand motion recovery can scale more effectively to large-scale embodied data generation pipelines.

\section{Embodied Foundation Models and Task-Decision Operators}

Embodied foundation models and task-decision operators constitute the high-level semantic understanding and decision-making layer of embodied intelligence systems. They integrate visual observations, language instructions, task context, and sometimes robot proprioceptive states to support instruction understanding, semantic grounding, task planning, action generation, and outcome prediction. According to their input-output forms and functional roles, existing methods can be broadly categorized into three paradigms: vision-language models, vision-language-action models, and world-model-based embodied models.

\subsection{Vision-Language Models}

Vision-language models (VLMs) can serve as multimodal semantic understanding operators in embodied intelligence systems. Their primary role is usually not to directly control robots, but to understand visual and textual inputs and provide high-level semantic information for downstream planning modules, control policies, or task-execution systems. Given images, videos, and language instructions, VLMs can support scene description, visual question answering, object recognition, semantic grounding, text recognition, and task-relevant contextual understanding.

Qwen-VL \cite{Qwen-VL} is a representative family of multimodal large models. Built upon the Qwen language model and trained on large-scale image-text data, this model family demonstrates strong capabilities in image understanding, visual question answering, and fine-grained visual semantic parsing. For embodied intelligence systems, models such as Qwen-VL can serve as semantic perception and instruction-grounding modules, aligning natural-language task descriptions with objects, regions, and relations in visual scenes, thereby providing high-level semantic priors for subsequent task planning and action execution.

However, VLMs themselves typically lack the ability to directly generate low-level robotic actions. They are better suited as an `understanding layer'' or `semantic interface'', responsible for answering questions such as `what is in the current scene'', `what is the task goal'', and ``which object or region should be attended to''. Concrete motion trajectories, control commands, and closed-loop execution still need to be handled by policy models, planners, or controllers.

\subsection{Vision-Language-Action Models}

Vision-language-action models (VLAs) further extend multimodal understanding capabilities to robotic action generation. Unlike VLMs, which mainly output text or semantic results, VLA models usually take images, language instructions, and robot states as inputs, and directly output robotic actions, end-effector trajectories, or discrete control commands. Therefore, VLAs can be regarded as key operators that bridge semantic understanding and embodied execution.

Representative VLA methods include RT-2 \cite{zitkovich2023rt}, OpenVLA \cite{kim24openvla}, and the $\pi$  \cite{black2024pi_0} family of models. OpenVLA is a representative open-source VLA method. Its core idea is to formulate robotic actions as discrete tokens that can be predicted by a language model, rather than designing an additional complex continuous-control head. Specifically, OpenVLA discretizes each dimension of continuous robotic actions into a set of bins and maps them to special action tokens in the LLM vocabulary. The model then autoregressively predicts action tokens using the standard next-token prediction objective. Therefore, the action head of OpenVLA is essentially a discrete token classification head based on the LLM vocabulary: it outputs discretized action sequences, which are then dequantized back into continuous robotic control values. The advantage of this design is its structural simplicity and its ability to reuse the training and inference paradigm of language models. However, its limitations lie in the potential loss of fine-grained control precision caused by action discretization, as well as additional inference latency when multiple action dimensions are generated autoregressively.

In contrast to OpenVLA's discrete action-token formulation, the $\pi$ family places greater emphasis on action generation for continuous control. Taking $\pi_0$ as an example, it introduces a dedicated action expert on top of a pretrained vision-language model and adopts flow matching to generate continuous action distributions. Instead of simply discretizing each action dimension into tokens, this action head predicts continuous action chunks from noise through a generation process similar to diffusion models, thereby outputting a segment of continuous control commands at once. Since robotic control often requires high-frequency, smooth, and continuous actions, the flow-matching action head of $\pi_0$ is better suited for dexterous manipulation, bimanual coordination, and long-horizon continuous control scenarios.

In addition, recent VLA studies such as JoyAI-RA 0.1 \cite{zhang2026joyaira01foundationmodel} and Qwen-RobotManip \cite{qwenrobotmanip} further explore data construction, model architectures, and training paradigms for VLAs. JoyAI-RA 0.1 emphasizes a multi-source and multi-level pretraining framework that integrates web data, egocentric human manipulation videos, simulation trajectories, and real-robot data to improve generalization across scenes and robot embodiments. Qwen-RobotManip builds a general robotic manipulation model based on Qwen-VL and performs unified alignment of heterogeneous manipulation data at the levels of representation, motion, and behavior, enabling large-scale multi-source training to serve robotic manipulation tasks more stably. These works indicate that VLAs are evolving from data-driven policies for single robots and single tasks toward large-scale, multimodal, multitask, and cross-embodiment embodied foundation models.

Overall, the core differences among VLA methods are largely reflected in their action representations and action-head designs. OpenVLA adopts a discrete action-token classification head, which facilitates the reuse of the autoregressive training framework of language models. $\pi_0$ adopts a flow-matching continuous action head, which is more suitable for smooth and high-frequency continuous control. Recent methods such as JoyAI-RA 0.1 and Qwen-RobotManip further focus on multi-source data alignment, cross-embodiment generalization, and large-scale training paradigms. The future development of VLAs will depend not only on model scale, but also on action-space modeling, robot data quality, and cross-platform transferability.

\subsection{World Models}

World models focus on predicting future states and the consequences of actions. They do not merely understand the current scene or directly generate actions; rather, they learn environmental dynamics and assist planning by ``imagining'' future scenarios. Therefore, world models can be viewed as predictive imagination and planning operators.

LingBot-VA \cite{lingbot-va2026} is a representative embodied model based on world modeling. It formulates robot control as a causal world-modeling problem and uses an autoregressive diffusion framework to jointly learn future-frame prediction and policy execution. Its design includes shared visual-action latent representations, closed-loop scene modeling with environmental feedback, and an asynchronous inference mechanism for efficient control. In the embodied pipeline, world models can support long-horizon planning, action-outcome estimation, and data-efficient policy learning, enabling agents to predict possible future states before executing actions.

Recent Cosmos 3 \cite{nvidia2026cosmos3omnimodalworld} further reflects the trend of world models toward general-purpose foundation models for physical intelligence. Cosmos 3 is positioned as an open omnimodal world model for Physical AI, supporting unified modeling of text, images, videos, ambient audio, and action sequences. Unlike traditional world models that mainly perform video prediction, Cosmos 3 incorporates physical reasoning, world generation, and action generation into a unified framework. It can generate physically plausible future scenes from the current state and model subsequent action sequences. For embodied intelligence, it can be used not only for synthetic data generation and simulation environment construction, but also to help robots evaluate action consequences before execution and understand motion causality and spatial interaction relationships, thereby supporting policy learning, model-based planning, and the development of Physical AI policy models.

Recent studies have also explored using world models as environment simulators for VLA systems \cite{zhu2025wmpoworldmodelbasedpolicy,jiang2026wovrworldmodelsreliable,gao2026swordstylerobustworldmodels}. In this setting, world models provide a predictive rollout space for evaluating candidate VLA actions before real execution, showing their potential to improve action selection, reduce unsafe trials, and support long-horizon decision-making.

Compared with VLA models, world models place greater emphasis on prediction and simulation rather than directly mapping current observations to actions. VLAs are more suitable for immediate action decision-making, while world models are more suitable for long-horizon planning, counterfactual reasoning, and action-outcome evaluation. The two can form a complementary relationship: VLAs generate candidate actions or policies, while world models predict the future states that these actions may lead to and help select safer and more effective execution plans.

\subsection{Limitations}

Although VLAs and world models provide a unified modeling framework for embodied intelligence systems, spanning semantic understanding to action generation and current perception to future prediction, their application in real robotic systems still faces multiple challenges.

First, data scale and data quality remain core bottlenecks. VLAs require large amounts of high-quality manipulation data covering diverse tasks, scenes, objects, and robot embodiments, while real-robot data collection is costly, slow, and often constrained by hardware platforms. Existing datasets are still insufficient in terms of task diversity, interaction complexity, and cross-embodiment coverage, leading to limited generalization to new scenes, new objects, or new robot platforms. World models also rely on large-scale temporal data to learn reliable environmental dynamics, but physical interactions, contact changes, and long-term causal relationships in real scenes are often difficult to model sufficiently from limited data.

Second, the reliability of action generation and world prediction still needs improvement. Although VLA models can directly generate actions from visual observations and language instructions, they may still suffer from action drift, incorrect grounding, or execution failures in long-horizon tasks, fine-grained manipulation, occluded scenes, and dynamic environments. World models can predict future states and action consequences, but their predictions may gradually accumulate errors over time. This issue is particularly prominent in scenarios involving complex contact, deformable objects, or multi-agent interactions, where generated future scenes may not strictly comply with real physical laws. Therefore, improving the stability of action generation and the physical consistency of future prediction remains a key challenge for current embodied foundation models.

In addition, real-time performance and safety are critical factors that constrain deployment. VLAs and world models typically have large parameter counts, high inference latency, and substantial memory overhead, making it difficult to directly satisfy the low-latency requirements of closed-loop control in real robotic systems. In real-world environments, robot actions must also satisfy constraints such as collision avoidance, force-control limits, joint limits, and task safety, while actions generated by end-to-end models do not naturally provide verifiable safety guarantees. Therefore, practical systems still need to integrate VLAs or world models with traditional planners, low-level controllers, safety constraints, and human supervision mechanisms.

Finally, the evaluation protocols for current VLAs and world models remain underdeveloped. Many methods are mainly evaluated on offline datasets, simulation environments, or limited real-world tasks, making it difficult to comprehensively reflect their long-term stability, cross-platform transferability, and real-task success rates in the open world. Future research needs to establish more unified, reproducible, and application-oriented evaluation protocols, while further exploring data scaling, model compression, physical-consistency modeling, safety-constraint integration, and task-adaptive reasoning.

\section{Planning, Control, and System Support Operators}

Planning, control, and system support operators form the execution layer of an embodied intelligence system. They receive task goals, object and robot states, geometric maps, action proposals, and safety constraints, and convert them into grasp configurations, collision-free trajectories, controller references, communication messages, and runtime scheduling decisions. Unlike perception or task-decision operators, these operators must produce outputs that are not only semantically reasonable, but also kinematically feasible, dynamically executable, temporally synchronized, and compatible with the target robot and computing platform. Therefore, they provide the main interface between high-level intelligence and physical execution.

\subsection{Grasp Planning Operators}

Grasp planning operators transform object geometry, target identity, and scene context into executable grasp candidates. For a parallel-jaw gripper, a typical output includes a 6D gripper pose, gripper width, approach direction, confidence score, and optional contact points. For dexterous hands, the output may additionally include finger joint configurations, contact assignments, and force-closure-related scores. A complete grasp operator generally contains candidate generation, grasp-quality estimation, reachability filtering, collision checking, and approach-and-retreat trajectory construction.

Recent learning-based methods have improved grasp generation in cluttered and open-object settings. Contact-GraspNet directly predicts distributions of 6D parallel-jaw grasps from scene depth observations, reducing dependence on multi-stage candidate-generation pipelines \cite{sundermeyer2021contactgraspnet}. AnyGrasp further emphasizes dense, full-DoF, and temporally smooth grasp perception, making it suitable for continuous robot operation under viewpoint changes and depth noise \cite{fang2023anygrasp}. However, the output of a learned grasp model should not be treated as an immediately executable command. The grasp pose must still be transformed into the robot base frame, checked for inverse-kinematics feasibility and collision, and associated with task-specific approach, closure, lifting, and recovery behaviors. This distinction is important because a geometrically plausible grasp may still be unreachable, unsafe, or unsuitable for the subsequent manipulation objective.

\subsection{Trajectory Planning, Collision Checking, and Control}

Trajectory planning operators convert an initial robot state and a target configuration into a collision-free path or a time-parameterized trajectory. Their inputs may include joint states, end-effector goals, robot kinematics, joint and velocity limits, scene geometry, and task constraints. Their outputs typically include joint-space waypoints, timestamps, velocities, accelerations, and planning-status or failure codes. Sampling-based planning remains useful for global exploration, whereas trajectory optimization is effective for producing smooth motions under geometric and dynamic constraints. In practical systems, hybrid pipelines often combine inverse kinematics, geometric search, trajectory optimization, and online replanning.

MoveIt~2 provides a modular manipulation framework for ROS~2, integrating robot kinematics, planning scenes, plugin-based motion-planning pipelines, trajectory processing, and controller-based trajectory execution \cite{moveit2docs}. cuRobo is a representative recent system that accelerates inverse kinematics, collision checking, geometric planning, and trajectory optimization through GPU-parallel computation \cite{sundaralingam2023curobo}. These systems illustrate that planning and collision checking are increasingly implemented as tightly coupled operators rather than isolated algorithms. A collision-checking operator should support self-collision and robot--environment collision, discrete or continuous trajectory validation, and geometric representations such as meshes, primitive shapes, signed-distance fields, or depth-derived obstacle models. When possible, it should return minimum distance, collision pairs, gradients, and the violating waypoint or time, rather than only a binary valid--invalid result.

Control operators execute planned motion at a higher frequency than task planning. They translate reference trajectories into joint positions, velocities, torques, Cartesian impedance targets, or force-control commands while enforcing joint, speed, acceleration, torque, and contact-force limits. In closed-loop systems, perception and execution feedback may invalidate a grasp or trajectory, and the planning operator must then replan or activate a fallback behavior. This separation allows VLA or world-model outputs to provide task-level action proposals while deterministic planning and control operators enforce embodiment-specific feasibility and safety constraints.

\subsection{ROS 2 Adaptation and Data Transfer}

ROS~2 adaptation operators provide the communication and software-integration layer for planning and control modules. ROS~2 supports distributed robotic systems through nodes, topics, services, actions, executors, and configurable quality-of-service policies \cite{macenski2022ros2}. Within an embodied-operator library, adaptation is not limited to wrapping a model as a ROS node. It includes converting model-native tensors and arrays into robot messages, preserving coordinate-frame semantics, attaching timestamps and confidence values, synchronizing multi-sensor streams, configuring transport policies, and exposing long-running planning or manipulation procedures through action interfaces.

Data transfer is particularly important for high-bandwidth embodied workloads. Images, point clouds, depth maps, masks, and trajectories may otherwise undergo repeated serialization, host--device copies, and inter-process transfers. ROS~2 node composition can reduce communication overhead by colocating suitable components in one process \cite{macenski2023composition}, while Isaac ROS NITROS uses type adaptation and negotiation to reduce redundant data movement in accelerated ROS~2 graphs \cite{isaacrosnitros}. However, reduced-copy execution must preserve message ownership, timestamp consistency, coordinate metadata, and failure isolation. A locally faster data path is not useful if it introduces unsafe buffer reuse, hidden synchronization, or backend-specific interfaces that prevent portability.

\subsection{Heterogeneous Scheduling and Runtime Coordination}

Heterogeneous scheduling operators coordinate workloads across CPUs, GPUs, NPUs, FPGAs, and robot-side microcontrollers. Embodied systems contain computational chains with different requirements: high-rate control and safety callbacks require bounded response time, perception and VLA inference require accelerator access, and mapping or logging tasks may tolerate delay but consume substantial memory and bandwidth. A useful scheduling operator should therefore consider priority, deadline, device affinity, memory availability, stream concurrency, preemption capability, and communication cost rather than optimizing average utilization alone.

RobotCore proposes a target- and accelerator-agnostic architecture for integrating hardware acceleration into ROS~2 computational graphs \cite{mayoral2022robotcore}. PAAM further studies coordinated and priority-driven access to shared accelerators for time- and safety-critical ROS~2 callback chains \cite{enright2024paam}. RobotPerf provides a vendor-agnostic framework for evaluating robotics computing graphs across heterogeneous hardware platforms \cite{mayoral2024robotperf}. Together, these works show that scheduling is part of robot functionality rather than a transparent backend detail: scheduling decisions directly affect sensing freshness, planning deadlines, controller jitter, and ultimately task safety.

\subsection{Limitations}

Planning, control, and system support operators still face several open challenges. Learned grasp and motion generators may produce geometrically plausible but task-inappropriate actions, whereas deterministic planners may fail when scene geometry is incomplete or rapidly changing. Collision models can become stale because of perception latency, and high-level plans may violate contact, force, balance, or timing constraints that are not represented in the planning state. At the system level, inconsistent coordinate frames, timestamp drift, message conversion, memory copies, executor contention, and unpredictable accelerator sharing can dominate end-to-end latency even when individual models are fast.

Future operator libraries should therefore provide unified contracts for robot models, coordinate frames, timing, uncertainty, constraints, failure codes, and fallback behaviors. They should support incremental scene updates, online replanning, traceable resource scheduling, and portable execution across heterogeneous hardware. Most importantly, these operators should be evaluated as a connected execution pipeline: a useful system must transform perception and decision outputs into safe robot behavior with predictable latency and recoverable failure modes. These requirements motivate the multi-dimensional embodied-operator benchmark introduced in the next section.

\section{Embodied Operator Benchmark}

\subsection{Motivation and Scope}

Embodied operators require a benchmark that differs from conventional single-model evaluation. A model benchmark usually measures the predictive accuracy or inference latency of a fixed neural network under a predefined input format. In contrast, an embodied-operator benchmark should evaluate how a functional module behaves as a deployable component within a broader embodied workflow. Such a benchmark must therefore consider not only output correctness and runtime efficiency, but also temporal stability, resource usage, cross-platform reproducibility, interface compatibility, and downstream task utility. These dimensions are necessary because an operator may achieve strong standalone performance while still failing to provide reliable support for long-video processing, robot learning, online perception-control loops, or real-world deployment.

The proposed benchmark organizes embodied computation into a hierarchical evaluation scope. At the lower level, it examines computational efficiency and graph execution behavior. At the functional level, it evaluates whether an operator satisfies its task semantics and preserves the required input-output contract. At the robot workflow level, it further measures whether the operator can support planning, control, interaction, or data generation in a stable manner. At the task and platform levels, the benchmark considers whether multiple operators can be integrated into complete embodied applications with reproducible performance across hardware platforms, software environments, and deployment settings. This work mainly focuses on the functional operator level, while defining metrics that can be extended toward robot workflows and compound embodied tasks. This system-level scope is consistent with recent embodied-AI surveys, which jointly discuss perception, interaction, robot data, simulation, and downstream task evaluation \cite{liu2025aligning}.

\subsection{Benchmark Protocol}

Each benchmarked operator should be registered with an explicit input-output contract. For a segmentation operator, the input contract includes image or video format, prompt type, frame resolution, and camera metadata when available; the output contract includes masks, labels, track identifiers, confidence scores, and serialization format. For a hand motion recovery operator, the input contract includes video, camera parameters, optional stereo information, and hand-detection priors; the output contract includes MANO parameters, global hand trajectories, temporal correspondences, confidence values, and failure flags. Such contracts prevent ambiguous comparisons in which different systems evaluate different tasks under the same operator name.

The scene suite should include short diagnostic cases, medium-scale production cases, and long stress cases. Diagnostic cases isolate model behavior and facilitate correctness checks. Medium cases reflect typical deployment workloads. Long stress cases reveal memory leaks, cache behavior, scheduling instability, I/O bottlenecks, and accumulated numerical drift. For hand-object segmentation, the scene suite should include small objects, large objects, heavy occlusion, fast hand motion, reflective surfaces, object transfer between hands, and background clutter. For hand motion recovery, it should include static and moving cameras, one-hand and two-hand interactions, self-occlusion, missing detections, rapid motion blur, camera scale ambiguity, world-coordinate trajectory drift, and out-of-view hand motion. DexYCB provides a useful reference for constructing grasping-oriented hand-object benchmark cases because it evaluates 2D object and keypoint detection, 6D object pose estimation, 3D hand pose estimation, and safe human-to-robot handover under hand grasping scenarios \cite{chao2021dexycb}. HaWoR further suggests that egocentric hand-motion benchmarks should consider world-space hand trajectories, camera trajectory estimation, and missing-frame completion, rather than only camera-frame hand pose accuracy \cite{zhang2025haworworldspacehandmotion}. Recent egocentric hand-object datasets provide additional references for constructing such scene diversity, especially when hand trajectories, object poses, and interaction states need to be evaluated jointly \cite{chao2021dexycb,banerjee2025hot3d}.

The benchmark should report accuracy, latency, resource, stability, and portability metrics jointly. Accuracy metrics depend on the operator category. Segmentation should report mask IoU, boundary F1, track consistency, object-presence precision and recall, and downstream crop validity. Hand motion recovery should report 2D reprojection error, 3D trajectory error, threshold success rates, temporal smoothness, hand-interaction plausibility, and failure-case statistics. Latency metrics should include end-to-end time, stage-level time, median and tail latency, throughput, cold-start time, cache-build time, and output serialization time. Resource metrics should include peak GPU memory, CPU memory, GPU utilization, CPU utilization, memory-bandwidth sensitivity, disk I/O, and host-device transfer volume. Stability metrics should include output determinism, variance across repeated runs, long-video failure rate, and sensitivity to missing frames or prompts. Portability metrics should include performance across container images, driver versions, thread settings, GPU types, CPU-only variants, TensorRT or OpenVINO backends, and shared versus exclusive queues.

\subsection{Benchmark Registration and Reporting Rules}

To ensure reproducibility in embodied-operator evaluation, every benchmark run must be accompanied by an Operator Card and a Run Manifest. The Operator Card shall record the operator name, functional category, input-output contract, model checkpoint, pre- and post-processing versions, supported backends, and known failure modes. The Run Manifest shall record the dataset split, scenario list, hardware platform, driver and runtime versions, container image, thread settings, batch configuration, random seed, precision mode, caching strategy, and evaluation script version. This metadata is indispensable because embodied operators are typically composite systems rather than single neural networks; small differences in data transformation, serialization, backend selection, or runtime scheduling can lead to significant end-to-end behavioral changes.

Benchmarks shall partition evaluation into the following three tracks:

\begin{itemize}
    \item \textbf{Correctness-Preserving Track}: This track is designed for optimizations expected to preserve outputs exactly or maintain task equivalence, such as cache reuse, I/O elimination, memory management, online scheduling, and serialization optimization.
    \item \textbf{Approximate Output Track}: This track is designed for optimizations that introduce numerical differences, including backend conversion, precision reduction, quantization, TensorRT/OpenVINO deployment, and algorithmic approximation. It must report both output-layer discrepancies and task-level accuracy changes.
    \item \textbf{Deployment Track}: This track focuses on whether an operator can function as a stable component in a real-world system, with emphasis on end-to-end latency, resource footprint, portability, fault recovery, and downstream task impact.
\end{itemize}

Benchmarks shall also establish explicit metric aggregation rules. For short diagnostic cases, results should be reported per scenario to expose specific failure modes. For medium and long cases, metrics should be aggregated using both frame-level and scenario-level statistics. For accuracy, the failure rate and threshold success rate should be reported in addition to the mean value. For latency, the median, P95, and P99 percentiles should be reported, with a distinction between cold-start latency and steady-state latency. Resource metrics should cover both peak and time-averaged utilization. When videos differ in the number of hands, object instances, trajectories, or valid frames, the benchmark must clearly state whether aggregated results are weighted by frame count, instance count, scenario count, or task count.

Beyond standard clean inputs, benchmarks should incorporate perturbation and stress testing. Perturbation tests may inject anomalies such as missing frames, dropped prompts, corrupted masks, noisy camera parameters, resolution changes, timestamp misalignment, and intermittent detection failures. Stress tests should cover long-video processing, repeated execution, multi-stream input, shared-device execution, and memory-constrained scenarios. Such tests are essential because embodied operators are typically deployed in long-running robotic systems or data production pipelines; their robustness, determinism, and recovery behavior are therefore as important as single-run accuracy.

Finally, benchmark reports must include regression criteria. An optimization should be considered valid only if it improves at least one target dimension without causing unacceptable degradation beyond predefined tolerance in any other dimension. For example, a speed optimization must report whether accuracy, determinism, failure rate, peak memory, and downstream task success rate remain within predefined tolerances. This mechanism prevents an optimization from being accepted solely because it improves a single metric, such as FPS or model inference latency.

\subsection{Benchmark Metrics for the Studied Operators}

Table~\ref{tab:operator_metrics} lists the primary evaluation metrics by operator category. These metrics span the core dimensions of correctness, efficiency, and stability, and directly correspond to the measurement objectives defined in Table~\ref{tab:benchmark}.

\begin{table}[h]
\centering
\caption{Benchmark metrics for the studied operator categories.}
\label{tab:operator_metrics}
\begin{tabular}{L{0.28\linewidth}L{0.62\linewidth}}
\toprule
\textbf{Operator Category} & \textbf{Representative Metrics} \\
\midrule
Detection and Segmentation Operators & mAP, AP$_{50}$, AP$_{75}$, mIoU, IoU, F1, Pixel Accuracy, Boundary IoU, Mask Stability \\
\midrule
Spatial Localization and 3D Understanding & ATE, RTE, RPE, Reprojection Error, Abs Rel, $\delta_{<1.25}$, ADD, ADD-S, 3D IoU, Chamfer Distance \\
\midrule
Hand Motion Recovery Operators & MPJPE, PA-MPJPE, N-MPJPE, PCK@$\alpha$, AUC, Hand mIoU, Mesh Vertex Error, F-score \\
\midrule
Embodied Foundation Models and Task-Decision Operators & Task Success Rate, Sub-goal Completion, Action Accuracy, Plan Validity, Instruction Following Accuracy, Decision Latency \\
\midrule
Planning, Control, and System Support Operators & Trajectory Error (RMSE), Collision Rate, Tracking Error, Settling Time, E2E Latency, P95/P99 Latency, Peak Memory, Throughput, Recovery Time \\
\bottomrule
\end{tabular}
\end{table}

\paragraph{Reporting Specifications}
For detection and segmentation operators, mAP and mIoU are the core metrics; for hand recovery operators, MPJPE and PCK are primary metrics; for 3D understanding operators, ATE/RPE and ADD-S are primary metrics. Decision operators must report task success rate and decision latency. Planning and control operators must distinguish between cold-start and steady-state latency, and report P95/P99 tail latency and peak memory consumption. All categories should additionally report failure rate and recovery time under stress testing.

\begin{table}[t]
\centering
\caption{Proposed embodied-operator benchmark dimensions.}
\label{tab:benchmark}
\begin{tabular}{L{0.20\linewidth}L{0.32\linewidth}L{0.34\linewidth}}
\toprule
Dimension & Measurement Target & Example Metrics \\
\midrule
Correctness & Whether the operator satisfies its task contract & IoU, F1, reprojection error, trajectory error, failure rate \\
\midrule
Efficiency & Whether the operator is fast enough end to end & Total latency, stage latency, throughput, tail latency, cold start \\
\midrule
Resource Use & Whether the operator uses hardware efficiently & GPU memory, CPU memory, utilization, I/O, host-device transfer \\
\midrule
Stability & Whether outputs and latency are reproducible & Repeated-run variance, long-video robustness, deterministic hashes \\
\midrule
Portability & Whether results transfer across environments & Driver/image sensitivity, CPU/GPU variants, backend compatibility \\
\midrule
Task Utility & Whether downstream embodied tasks benefit & Grasp success, data validity, retargeting quality, manual correction rate \\
\bottomrule
\end{tabular}
\end{table}

\subsection{Operator Acceleration and Benchmark Design}

Operator acceleration is a key requirement for embodied intelligence, but it should not be understood as neural network inference acceleration alone. Embodied operators are usually executed inside perception-action loops, long-video processing pipelines, robot data workflows, simulation environments, and deployment systems. Their efficiency is jointly affected by model computation, sensor streaming, memory transfer, temporal state update, runtime scheduling, output serialization, and safety fallback. Therefore, acceleration should be evaluated at the operator level, where the input-output interface, downstream utility, and safety constraints remain explicit.

A central principle is to distinguish local model speedup from workflow-level acceleration. For example, accelerating a segmentation decoder does not necessarily improve the complete video segmentation operator if visual encoding, prompt processing, memory update, or result serialization remains the bottleneck. Similarly, faster policy decoding may not improve robot execution if action post-processing, communication latency, or safety filtering dominates the control loop. Thus, benchmark design should report both internal stage latency and end-to-end operator latency.

Recent industrial and open-source systems provide practical references for embodied-operator acceleration. Isaac ROS NITROS reduces redundant data movement in ROS~2 perception graphs through type adaptation and type negotiation, which is important for pipelines that continuously exchange images, masks, poses, and geometric representations \cite{isaacrosnitros}. TensorRT supports graph optimization, mixed-precision inference, dynamic-shape execution, and quantized computation for hardware-aware neural deployment \cite{tensorrt}. DeepStream organizes decoding, inference, tracking, and output generation as a streaming video pipeline, which is relevant to multi-camera perception and long-video processing \cite{deepstream}. Isaac Sim and Isaac Lab extend acceleration to GPU-based simulation, rendering, sensor modeling, data generation, reinforcement learning, and imitation learning \cite{isaacsim,mittal2025isaac}. LeRobot further shows that robot communication, data collection, dataset management, policy training, and asynchronous inference can be integrated into a unified robot learning stack \cite{cadene2026lerobot}.

For embodied perception operators, acceleration should preserve geometric and temporal reliability. Video segmentation, hand tracking, stereo depth, pose estimation, and 3D scene understanding are not isolated image inference tasks. A SAM2-style segmentation operator depends on visual encoding, prompt-conditioned mask generation, temporal memory update, and mask propagation \cite{ravi2025sam2}. A Dyn-HaMR-style hand recovery operator further involves hand reconstruction, camera-aware optimization, temporal fitting, and trajectory estimation \cite{yu2025dyn}. Therefore, perception acceleration should be evaluated by end-to-end latency, memory-transfer cost, mask or pose accuracy, temporal consistency, and downstream manipulation validity. A faster operator should not be accepted if it increases mask drift, unstable pose estimation, tracking failure, or contact-boundary error.

For VLA and action operators, acceleration should focus on representation efficiency and closed-loop control quality. Unlike conventional vision-language inference, VLA outputs are executed by physical robots and are constrained by control frequency, motion continuity, and safety. FAST reduces action sequence length through frequency-domain action tokenization, showing that action representation itself can be an acceleration target \cite{pertsch2025fast}. OpenVLA and OpenVLA-OFT improve adaptation and action decoding efficiency through action token prediction, LoRA-based fine-tuning, parallel decoding, and action chunking \cite{kim24openvla,kim2025fine}. $\pi_0$ and GR00T N1 explore continuous or hierarchical action generation for general robot control \cite{black2024pi_0,bjorck2025gr00t}. Accordingly, VLA acceleration should be measured by token efficiency, action latency, task success rate, motion smoothness, recovery ability, and unsafe action rate, rather than by language-model throughput alone.

Data and simulation operators should also be included in acceleration benchmarks because they determine whether large-scale embodied learning can be reproduced efficiently. Large robot datasets such as Open X-Embodiment and DROID increase the importance of standardized data schemas, temporal alignment, episode replay, and efficient batch construction \cite{o2024open,khazatsky2024droid}. Slow training may originate from data decoding, CPU preprocessing, storage access, or inefficient sampling rather than from the model itself. Simulation operators provide another acceleration path by generating diverse tasks and interaction data at scale. However, simulation speed is useful only when physical plausibility, sensor fidelity, task coverage, and sim-to-real transfer are preserved. Therefore, data and simulation benchmarks should report loading throughput, timestamp alignment error, replay fidelity, simulation throughput, parallel environment count, and downstream training or transfer performance.

Deployment operators convert research models into executable modules on target hardware. Their acceleration includes model export, graph compilation, precision conversion, memory management, runtime scheduling, monitoring, and fallback handling. For embodied systems, average latency is insufficient because tail latency may destabilize closed-loop control. Benchmarks should therefore report P95/P99 latency, peak memory, power consumption when available, communication overhead, backend portability, and recovery time. Acceleration should not remove confidence estimation, runtime logging, failure flags, safety filtering, or fallback mechanisms, because these components are essential for reliable physical deployment.

Overall, acceleration claims for embodied operators should follow a consistent benchmark protocol. The operator interface should remain unchanged unless an approximate-output track is explicitly defined. Reported speedup should include both stage-level and end-to-end latency, together with resource usage, data movement, temporal consistency, downstream task utility, and safety behavior. An acceleration method should not be regarded as effective if it improves local runtime while degrading mask quality, pose stability, action reliability, replay fidelity, simulation realism, or closed-loop safety.

\subsection{Open Questions and Application Prospects for Embodied Operators}
\label{sec:open-questions-and-application-prospects}

This subsection summarizes the open questions facing embodied-operator platforms and translates them into actionable application and platform directions. We first motivate the importance of embodied operators, identify three structural tensions that such platforms must resolve, summarize four priority open questions, outline promising application scenarios, and propose a three-stage platform roadmap.

\subsubsection{Why Embodied Operators Matter}

In this work, an embodied operator denotes a reusable computational unit for real-world robotic tasks rather than a tensor-level primitive or an isolated model checkpoint. A typical operator couples multimodal perception, 3D spatial reasoning, robot-state processing, action generation, planning and control, runtime scheduling, and safety constraints, consistent with the system-level view emphasized in embodied-AI and edge-deployment surveys~\cite{liu2025aligning,grover2026embodied}. As embodied intelligence moves from laboratory demonstrations toward manufacturing, warehousing, inspection, healthcare, and household services, the value of such operators lies in reusable end-to-end task pipelines that can be adapted across hardware platforms, embodiments, and application contexts~\cite{o2024open,khazatsky2024droid}.

Three developments motivate this shift. First, robot foundation models are evolving toward VLA and embodied-reasoning systems: Gemini Robotics and Gemini Robotics-ER emphasize visuospatial understanding, task planning, progress estimation, and action grounding, while GR00T N-series and Cosmos connect simulation, synthetic data, policy training, evaluation, and edge deployment into integrated stacks~\cite{geminirobotics2025,bjorck2025gr00t,nvidia2025cosmos}. Second, robot-data standardization is accelerating through LeRobotDataset and cross-embodiment datasets, which organize multimodal time-series data, sensor signals, and multi-camera video into reusable formats~\cite{cadene2026lerobot,o2024open,wang2024all}. Third, humanoid and mobile-manipulation robots are increasingly evaluated in production and logistics settings, consistent with the growth trends reported for industrial and service robotics~\cite{ifr2025industrial,ifr2025service}.

\subsubsection{Core Tensions}

Embodied-operator platforms must resolve three coupled tensions. The first is the gap between general models and heterogeneous embodiments. VLA and world models aim to support broad physical intelligence, but robots differ in degrees of freedom, end effectors, sensors, control frequencies, and action spaces, which requires explicit embodiment-adaptation layers~\cite{zitkovich2023rt,kim24openvla,o2024open}. The second is the gap between module accuracy and end-to-end performance. Improving a segmentation, depth, pose, or VLA module does not necessarily improve task success when image acquisition, data movement, coordinate transforms, synchronization, control frequency, collision checking, and safety policies dominate the critical path~\cite{macenski2023composition,isaacrosnitros,mayoral2022robotcore,enright2024paam,mayoral2024robotperf,grover2026embodied}. The third is the gap between open-ended applications and verifiable deployment. Factories, warehouses, hospitals, and homes are not standardized benchmarks; therefore, task success, exception recovery, human intervention, maintenance cost, ROI, and safety events must be measurable and auditable~\cite{iso10218_1_2025,iso10218_2_2025,ifr2025industrial}. These tensions position embodied operators as an intermediate layer between foundation models, robot hardware, and industry workflows~\cite{liu2025aligning,li2026vision}.

\subsubsection{Open Questions}

The main open questions can be organized into four directions that are directly aligned with the benchmark requirements discussed above.

\textbf{Operator boundaries and composition.} If the granularity is too fine, an embodied operator reduces to a conventional model or kernel operator; if it is too coarse, important distinctions among detection, segmentation, depth, 6D pose, grasp planning, trajectory planning, and execution feedback become hidden. A practical design should therefore adopt a layered system spanning primitive compute operators, model-function operators, robot-function operators, composite-task operators, and platform-level operators. Each operator should specify its task semantics, input-output contract, supported hardware, validated scenarios, failure modes, and fallback behavior, so that execution constraints are explicit rather than implicitly embedded in model weights.

\textbf{Data standards, world models, and sim-to-real.} Embodied data are more heterogeneous than standard vision or language corpora, covering RGB streams, depth, camera calibration, point clouds, joint states, end-effector poses, torque, tactile signals, language, action sequences, control rates, success labels, failure cases, and human-intervention records~\cite{o2024open,khazatsky2024droid,cadene2026lerobot}. LeRobotDataset and other cross-embodiment efforts reduce file-level fragmentation, but unresolved issues remain in action-space alignment, timestamp synchronization, coordinate-frame conventions, failure-data coverage, and metadata completeness~\cite{cadene2026lerobot,wang2024all}. Because real-robot data collection is costly and safety-constrained, practical systems increasingly combine real data, simulation, and generative world models~\cite{nasiriany2024robocasa,nasiriany2026robocasa365,nvidia2025cosmos}. The central challenge is to operatorize simulation and synthetic-data generation while preserving physical plausibility, long-tail coverage, and real-task validation~\cite{isaacsim,mittal2025isaac}.

\textbf{VLA, safety, and edge deployment.} VLA and embodied-reasoning models should not be treated as unconstrained robot controllers. They may misunderstand instructions, generate dynamically infeasible actions, fail to maintain high-frequency control, or introduce physical safety risks~\cite{zhang2025safevla,li2026vision,qin2026embodiedgovbench}. A more defensible architecture is hierarchical. A low-frequency layer, such as VLM, VLA, Gemini Robotics-ER, or GR00T, performs language grounding, scene reasoning, and task decomposition~\cite{geminirobotics2025,bjorck2025gr00t,kim24openvla}. A mid-frequency layer, including segmentation, depth, 6D pose, and diffusion policy, supports tracking, pose estimation, and sub-action generation~\cite{ravi2025sam2,wen2024foundationpose,chi2023diffusion}. A high-frequency layer, such as MoveIt, Nav2, MPC, and safety controllers, executes trajectories, obstacle avoidance, force control, balance, and emergency stops~\cite{chitta2012moveit,macenski2020marathon2,mayne2000constrained,iso10218_1_2025}. In this design, VLA primarily acts as a task-decision operator, while deterministic planners and controllers enforce collision checking, motion constraints, velocity and torque limits, and safety shutdowns \cite{moveit2docs,macenski2022ros2,iso10218_1_2025,iso10218_2_2025,mayne2000constrained}.

\textbf{Evaluation and operational value.} The value of embodied intelligence is ultimately determined by whether robots can complete tasks reliably, safely, and with limited intervention over long horizons~\cite{liu2025aligning,qin2026embodiedgovbench}. Evaluation should therefore cover five levels: model accuracy, module latency and resource use, pipeline-level latency, task-level success and recovery, and operational metrics such as continuous runtime, intervention rate, maintenance cost, utilization, ROI, and safety events~\cite{grover2026embodied,ifr2025industrial}. A credible platform should connect operator-level improvements to operational value. In practical terms, lower miss rates and stronger occlusion handling can reduce grasp failures, more stable 6D pose estimation can improve assembly precision, lower end-to-end latency can increase cycle rate, and better exception recovery can reduce human intervention.

\subsubsection{Application Prospects}

Embodied operators are unlikely to diffuse uniformly. Near-term deployment is more plausible in scenarios with structural regularity, explicit safety boundaries, and observable ROI~\cite{ifr2025industrial,iso10218_2_2025}. Industrial manufacturing has the clearest near-term profile, while warehousing, logistics, inspection, and maintenance also show relatively high readiness but face medium-to-high complexity due to long-tail objects, mobile manipulation, navigation, auditability, and exception handling. Healthcare, agriculture, and household services offer long-term potential, but their near-term readiness remains lower because of environmental variability, human-robot interaction requirements, and stricter safety constraints~\cite{goodrich2007human,iso15066_2016,ifr2025service}. In parallel, embodied-data production and alternative edge-chip adaptation are better understood as enabling scenarios: they support the broader ecosystem but still require reliable data pipelines, model conversion, heterogeneous scheduling, and benchmark validation~\cite{cadene2026lerobot,grover2026embodied}.

\subsubsection{Platform Roadmap}

An embodied-operator platform can be developed in three stages. In the near term, the focus should be on standardized and deterministic operators with explicit input-output contracts, model packaging, profiling, baseline runtimes, and initial hardware adaptation~\cite{grover2026embodied,isaacrosnitros}. The priority set includes open-vocabulary detection and video
segmentation \cite{liu2024grounding,ravi2025sam2}, stereo depth and 6D pose estimation \cite{wen2024foundationpose,wen2025foundationstereo}, SLAM and navigation \cite{campos2021orb,macenski2020marathon2}, grasp and trajectory planning \cite{sundermeyer2021contactgraspnet,fang2023anygrasp,moveit2docs,sundaralingam2023curobo}, motion recovery \cite{zhang2025haworworldspacehandmotion,yu2025dyn}, and deployment, communication, and scheduling \cite{macenski2022ros2,macenski2023composition,mayoral2022robotcore,enright2024paam,mayoral2024robotperf,grover2026embodied}. In the medium term, the objective should shift from module reuse to task-pipeline reuse. Industrial pipelines may combine localization, segmentation, depth, pose, grasp planning, trajectory planning, execution, and quality feedback; inspection pipelines may combine navigation, OCR or VLM-based reading, anomaly assessment, report generation, and human review; warehouse pipelines may combine order parsing, item recognition, graspable-region generation, pallet planning, AMR scheduling, and inventory writeback~\cite{ifr2025industrial,ifr2025service}. In the long term, the platform should evolve into embodied-intelligence infrastructure covering data governance, model management, operator libraries, simulation and evaluation, deployment runtime, industry pipelines, and operational monitoring~\cite{liu2025aligning,cadene2026lerobot,nasiriany2026robocasa365}. Its central advantage lies in ecosystem-level capability: supporting multiple embodiments, adapting to multiple chip platforms, integrating with business systems, continuously converting execution data into model improvement, and providing explainable and maintainable task capabilities~\cite{o2024open,grover2026embodied}.

\subsubsection{Summary and Outlook}

The open questions surrounding embodied operators are closely tied to the broader transition of embodied intelligence from model capability to deployable infrastructure~\cite{liu2025aligning,grover2026embodied}. Strong point capabilities already exist in visual segmentation, depth estimation, pose estimation, SLAM, action understanding, VLA, planning, and control~\cite{ravi2025sam2,wen2025foundationstereo,wen2024foundationpose,campos2021orb,zitkovich2023rt,kim24openvla,chitta2012moveit,sun2026prevlapreemptiveruntimeverification}. The key deployment question, however, is whether these capabilities can be composed into systems that remain accurate, efficient, safe, and maintainable over long-horizon execution~\cite{qin2026embodiedgovbench,ifr2025industrial}. Large-scale infrastructure efforts, such as thousand-GPU training and optimization systems, also suggest that embodied-operator platforms will require scalable computation, distributed training, and system-level optimization support~\cite{guo2026thousandgpulargescaletrainingoptimization}. Future platforms should therefore emphasize layered operator abstractions, cross-embodiment data standards, simulation and world-model loops validated by real tasks, hierarchical interfaces between VLA reasoning and deterministic control, and task-level benchmarks that translate accuracy, latency, and resource use into success rate, intervention rate, deployment cost, and ROI~\cite{o2024open,cadene2026lerobot,nvidia2025cosmos,mittal2025isaac,li2026vision}. In application terms, industrial manufacturing, warehousing and logistics, inspection and maintenance, embodied-data production, and alternative edge-chip adaptation are the most plausible near-term directions, while healthcare, agriculture, outdoor operations, and household services remain important but more demanding long-term targets~\cite{ifr2025industrial,ifr2025service,goodrich2007human}.

\section{Conclusion}

This work defines embodied operators as reusable and deployable functional modules for embodied intelligence systems, and organizes them into five categories: detection and segmentation, spatial localization and 3D understanding, hand motion recovery, embodied foundation models and task decision, and planning, control, and system support. These operators connect perception, reasoning, learning, planning, and execution, and provide the intermediate representations and executable outputs required by practical robotic workflows.

We further argue that embodied operators should be evaluated as holistic system components rather than isolated neural networks. Therefore, their benchmark should consider not only accuracy, but also end-to-end latency, resource usage, temporal stability, interface compatibility, deployment reliability, failure recovery, and downstream task utility. Overall, this work provides a modular perspective for building reusable, scalable, and verifiable embodied intelligence systems.

\newpage

\section*{Contributions}

\noindent\textbf{Author List$^{*}$}\par
\vspace{0.4em}

\noindent
Junwu Xiong$^{1}$,
Wei Chai$^{1,4}$,
Renxing Chen$^{1}$,
Jiaxuan Gao$^{1,3}$,
Yu Guo$^{1,4}$,
Yucheng Guo$^{1}$,
Yuzhen Li$^{1}$,
Mingxi Luo$^{1}$,
Wenyang Ma$^{1,4}$,
Yiyun Mou$^{1,4}$,
Yifei Zhang$^{1,4}$,
Chen Zhou$^{1}$,
Yongjian Guo$^{1,2}$\par

\vspace{0.8em}
\noindent\textbf{Affiliations}\par
\vspace{0.35em}

\noindent
$^{1}$\,AI Infra Team at JDT\\
$^{2}$\,Tsinghua University\\
$^{3}$\,Tianjin University\\
$^{4}$\,Beihang University\\

\bibliographystyle{unsrt}
\bibliography{main}

\clearpage



\end{document}